\documentclass{article}

\PassOptionsToPackage{numbers}{natbib}
\usepackage[dblblindworkshop,final]{neurips_2026}
\usepackage{amsmath}
\usepackage{multirow}
\usepackage{amssymb}   
\workshoptitle{Foundations of Agentic Systems Theory (FAST)}

\usepackage[utf8]{inputenc} 
\usepackage[T1]{fontenc}    
\usepackage{hyperref}       
\usepackage{url}            
\usepackage{booktabs}       
\usepackage{amsfonts}       
\usepackage{nicefrac}       
\usepackage{microtype}      
\usepackage{xcolor}         
\usepackage{graphicx}
\usepackage{subcaption}

\title{Coherence-Driven Belief Formation and Population Dynamics of Contagion in LLM Agents} 

\author{%
  Tathagata Banerjee\thanks{Equal contribution.} \\
  Takeda Pharmaceuticals \\
  \texttt{tathagata.banerjee@takeda.com} \\
  \And
  Nima Moghaddas\footnotemark[1] \\
  Northeastern University \\
  \texttt{moghaddas.n@northeastern.edu} \\
}

\begin{document}

\maketitle

\begin{abstract}
Models of social contagion usually assume how individuals adopt beliefs and derive population behavior from it. We instead empirically measure belief adoption in language model agents, quantifying the probability an agent adopts a claim given how many peers endorse it. We find this adoption kernel to be sigmoid, a characteristic of complex contagion, with a threshold that is sensitive to three sources: the claim's plausibility, the source's reliability, and the agent's disposition. These three dimensions are well approximated by a single effective dimension which we propose can be understood as the coherence of the incoming belief with the LLM agent's prior beliefs.  Further, we observe a characteristic of complex contagion in the collective dynamics of belief adoption in a system of AI agents: further spread on clustered than random networks. These systems also exhibit a bifurcating cascade window, and self-sustaining hysteretic consensus which lead to consensus being far harder to remove than to establish.

\end{abstract}

\section{Introduction}
AI agents increasingly operate in multi-agent teams, conversing, exchanging information, and
reaching decisions collectively \citep{wu2023}. Their beliefs are shaped not only
by one another but by the surrounding online information ecosystem; a recent large-scale study estimates that roughly one third of newly published web text is now AI-generated or AI-assisted
\citep{spennemann2025}. When agents act on what other agents have concluded, erroneous or manipulated beliefs can propagate through the system, spreading directly between agents or indirectly through
AI-generated text, an issue that recent opinion-simulation studies have begun to
surface \citep{chuang2024}.

A substantial body of work has studied social contagion in human networks. Theory distinguishes
simple contagion, which spreads through a single interaction, from complex contagion, which
requires reinforcement from multiple peers \citep{centola2007}. Empirical studies document
peer effects on individual behavior, most notably in observational studies of health
behaviors spreading through social ties \citep{christakis2007}, and show that individuals
jointly weigh consistency with prior beliefs and perceived source reliability when adopting a
claim \citep{hovland1951, kuutila2024}. These dynamics have been modeled: from threshold and
epidemic-style contagion models \citep{granovetter1978, watts2002}, to opinion-averaging and
bounded-confidence frameworks \citep{degroot1974, deffuant2000, hegselmann2002}, to Bayesian
and confirmation-bias models of how priors and source reliability shape adoption
\citep{acemoglu2011, allahverdyan2014}.

AI agents differ fundamentally from the agents in these models, which follow predefined
update rules often grounded in an assumption of rationality \citep{chuang2024}. AI agents can
read, interpret, and reason about a claim's content and source before adopting it, and can be
conditioned to reproduce fine-grained, demographically realistic human responses
\citep{argyle2023}, making them candidates for more realistic behavioral models. This has
motivated a growing line of work using AI agents to simulate social behavior, including generative
agent societies with emergent coordination \citep{park2023} and multi-agent debate in which
agents revise answers after seeing others' reasoning \citep{du2023}. 

Here, we study the factors that influence whether an AI agent adopts a belief and how it
spreads through a network. Rather than assuming an adoption rule, we measure the probability
an agent adopts a claim as a function of how many peers endorse it. We show that this adoption
kernel is a sigmoidal threshold response, sensitive to three factors: the plausibility of the
claim, the disposition of the adopting agent, and the reliability of its source. Each acts by
modifying the coherence between the model's prior beliefs and the incoming claim, and the
three are well approximated by a single dimension, suggesting one underlying axis of
coherence along which the model assesses new claims. The collective dynamics of belief adoption in a system of AI agents exhibits a characteristic of complex contagion: it spreads further on clustered than random networks \citep{centola2010}. These systems also exhibit a bifurcating cascade window, and self-sustaining hysteretic consensus which lead to consensus being far harder to remove than to establish.

\section{Related Work}
\paragraph{Social contagion and social influence.}
Studies of social contagion have shown that phenomena such as health behaviors cluster and spread across social ties \citep{christakis2007}. To explain how such spreading works, a key theoretical distinction separates simple contagion, which a single neighbor can transmit, from complex contagion, which requires reinforcement from multiple neighbors \citep{centola2007}. Experimental evidence has supported this account demonstrating socially reinforced behaviors spread further on clustered networks than on random ones \citep{centola2010}. Recent work, however, shows that this clustered-network advantage is conditional. It emerges when adoption without reinforcement is very low and adoption with reinforcement is very high \citep{doi:10.1073/pnas.2422892122}. In parallel, research on persuasion has shown that perceived credibility, prior beliefs, and source expertise jointly determine whether a claim
is accepted, from foundational studies of source credibility \citep{hovland1951} to recent
factorial experiments finding consistency with prior belief and source expertise to be the
strongest predictors of perceived credibility online \citep{kuutila2024}. The literature
identifies four recurring factors: social reinforcement, claim credibility, source expertise, and
prior belief. We ask whether these same factors organize belief adoption in LLM agents.

\paragraph{Classical agent-based models of opinion dynamics.}
A parallel line of work formalizes belief change as agents updating a numerical opinion from
their neighbors. In the DeGroot model an agent's opinion is a weighted average of its
neighbors' \citep{degroot1974}, and bounded-confidence models add that agents ignore
neighbors who are too dissimilar \citep{deffuant2000, hegselmann2002}. Closest to our framing, a cognitive model of interacting beliefs \citep{aiyappa2024} yields both simple- and complex-contagion adoption curves as an emergent property rather than by assumption, tested on the same clustered-versus-random paradigm as \citep{centola2010}.
Although these models capture consensus, polarization, and fragmentation, they represent
updating as a fixed rule applied regardless of the content or source of the information. We
instead replace this rule with an LLM agent that reasons over a claim's content, source, and
persona, asking whether the same regularities emerge from a content-sensitive mechanism.

\paragraph{Belief updating in language models.}
The stated positions of LLMs are sensitive to social input in ways that resemble, but are not
identical to, human belief dynamics. RLHF-trained models often shift toward a user's opinion
regardless of its correctness, an effect known as sycophancy \citep{perez2022, sharma2023}.
More directly, networked agents have been shown to update beliefs as a sigmoidal, threshold-like
function of the number of disagreeing neighbors, reminiscent of complex contagion, however they did not study how this varies with credibility, expertise, or clustering \citep{mehdizadeh2025}. Other work has found that assigning personas to agent communities changes their resilience to injected
misinformation depending on prior belief and source reliability \citep{lin2026},
\citep{savcisens2026}. We differ by bringing all of these factors into a single factorial
design, and by studying their effect on the adoption probability and the population level belief dynamics.

\section{Method}
\label{sec:method}

\subsection{Agents and the adoption kernel}
\label{sec:method-kernel}

An agent was a language model conditioned on a system prompt (its \emph{persona}) and a user
prompt which presented a claim and a summary of the opinions of its peers. For example, for a claim endorsed by $k$ of $n{=}10$
peers, the prompt stated ``$k$ of your 10 peers have already accepted this as true; the
other $10-k$ have not weighed in''. Finally, the agent was prompted to give its own belief by answering with a single word, \texttt{ADOPT} or
\texttt{DECLINE}. Unless stated otherwise we used a canonical peer-sensitive but non-credulous persona (full prompt templates in App.~\ref{app:repro}).

We measured the probability of adoption directly from the logits $\ell_{\mathrm{A}}, \ell_{\mathrm{D}}$ of the first sub-word tokens
of the two decisions and calculated $p_{\mathrm{adopt}}(k) = \sigma(\ell_{\mathrm{A}} -
\ell_{\mathrm{D}})$. Sweeping $k\in\{0,\dots,10\}$ yielded the \emph{adoption kernel}. For further analysis, we summarized the kernel by its threshold $\theta$, the linearly interpolated $k$ at which $p_{\mathrm{adopt}}$ crosses $0.5$ (defining $\theta{=}11$ if it never crossed, $\theta{=}0$ if it began above). 

\subsection{Beliefs and coherence manipulations}
\label{sec:method-substrate}
We considered belief statements about next-day Boston temperatures of the form ``Tomorrow it
will be $T^\circ$C in Boston.'' Although simple, this type of claim had three advantages: it
was (1) an unverifiable prediction the model could not resolve from stored knowledge, (2) free of
moral or ethical content that could have swayed the model, and (3) easy to make more or less
plausible by varying $T$. We replicated the results on a second claim type: next-day commute
times (App.~\ref{app:traffic}).

We manipulated the coherence between the incoming belief and the model's prior beliefs along
three axes. First, we varied the plausibility of the claim by changing $T$. Second, we varied the
reliability of the source, from a domain expert to an unreliable source. Third, we varied the
disposition of the focal agent, from credulous (highly socially attuned and trusting) to
hardened (highly skeptical and independent). See App.~\ref{app:repro} for prompting details.

To test whether these three axes acted by modifying the coherence between the incoming belief
and the model's priors, we installed an explicit prior belief in context and measured how the
\emph{adoption threshold} shifted. The adoption threshold is defined as the number of endorsing peers required to raise the adoption probability to 50\%. A \emph{claim prior} installed a belief relevant to the claim (e.g.,
an ongoing heat-wave vs.\ a cold-snap), altering the claim's coherence while the claim itself
is unchanged; a \emph{source prior} installed a belief about the source's track record; and a
\emph{disposition prior} installed a belief about how strongly the agent weights its own priors
against social input.

\subsection{Low-dimensional approximation}
\label{sec:method-reduction}

We tested whether these three axes of coherence (claim plausibility, source reliability, and
agent disposition) could be reduced to a single effective dimension. To do this, we measured the full
$3{\times}3{\times}3$ grid of claim$\times$source$\times$persona kernels, yielding a threshold
tensor $\Theta\in\mathbb{R}^{3\times3\times3}$, and analyzed it two ways. (1)
\emph{Dimensionality:} we unfolded $\Theta$ along its claim mode into a $3\times9$ matrix
(rows: the three claim levels; columns: the nine source$\times$persona combinations),
mean-centered each row, and took its singular value decomposition, recording the variance
captured by each principal component. (2) \emph{Additivity:} we fitted a main-effects model
$\hat\theta_{ijk}=\bar\theta+\alpha_i+\beta_j+\gamma_k$ (no interaction terms) and recorded its
$R^2$. 

\subsection{Population model} 
\label{sec:method-population}

We then asked what collective behavior emerged when multiple agents, each governed
by the measured kernel, influenced one another. We placed $N{=}30$ agents on a network and seeded
a fraction $\rho_0$ as initial believers, leaving the rest undecided, and used the kernel as the
local update rule. At each step, an undecided agent $i$ computed the fraction of its neighbors
that believed and expressed it as a count out of ten peers, $k_i$, and the agent then adopted with
probability $p_{\mathrm{adopt}}(k_i)$ from a live forward pass. Because each decision was an
actual model call rather than an averaged response, this was an agent-driven
simulation. We took the final believing fraction as the cascade outcome. 

For the clustering experiments we tested a characteristic of complex contagion: that contagion spreads further on clustered networks than on random ones
of equal density \citep{centola2010}. We generated Watts-Strogatz networks
of fixed mean degree and vary only the rewiring probability $p$. At $p{=}0$ the network is a
clustered ring lattice (high clustering coefficient $C$); at $p{=}1$ essentially random (low
$C$). We statistically compared the cascade size between these two configurations.

For the hysteresis experiments we tested whether a consensus, once
reached, sustained itself after the pressure that created it is withdrawn. This requires agents
that can lose a belief as well as gain one, so we measure a second \emph{retention}
kernel $p_{\mathrm{retain}}(k)$ using the analogous prompt for an agent that already holds the
belief and a \texttt{MAINTAIN}/\texttt{ABANDON} decision. Undecided agents adopt with
$p_{\mathrm{adopt}}(k_i)$ as before, while current believers abandon with $1-p_{\mathrm{retain}}(k_i)$. A fraction of agents are permanent zealots, who never abandon;ramping the zealot fraction up and then back down applies a controlled external pressure, and
comparing the up- and down-paths traces the hysteresis loop.

\subsection{Models}
\label{sec:method-stats}

Our primary model is \texttt{Qwen2.5-3B-Instruct} \citep{qwen2025}, with the pipeline
replicated on \texttt{Mistral-7B-Instruct-v0.3} \citep{jiang2023mistral}
(App.~\ref{app:mistral}). Full prompts, per-model tokens, plausibility mappings, network
parameters, and replicate counts appear in App.~\ref{app:repro}.\footnote{Code:
\url{https://anonymous.4open.science/r/multi-agent-llm-contagion-F402/}.}

\section{Individual belief adoption}
\label{sec:partI}
We first characterized how a single agent adopts a belief under social pressure. The adoption
kernels were sigmoidal, and were sensitive to three axes, each modifiable in a graded way. Further, the variation in adoption thresholds due to these axes were well approximated by a single effective dimension.

\subsection{Belief adoption is driven by coherence}
More plausible claims required less social reinforcement to be adopted, though even implausible
claims were adopted under sufficient social pressure (Fig.~\ref{fig1}a). Source reliability
shifted the threshold independently of the claim: a plausible claim that needed only $3$
endorsing expert peers required $9.4$ unreliable peers (interpolated threshold) before being
accepted (Fig.~\ref{fig1}b). Agent persona had a strong effect: credulous agents adopted both plausible and implausible
claims from a single peer, while hardened and skeptical agents were \emph{immune} (never adopted regardless of plausibility or social reinforcement), and only trusting agents were sensitive to the claim's plausibility (Fig.~\ref{fig1}c,d). Together, these show that an agent weighs attributes of the
belief, of its source, and perception of itself when deciding whether to adopt.

Next we sought to demonstrate whether the influence from these attributes were fixed or whether their effect was altered by an agents prior beliefs. We found strong evidence that prior beliefs affect adoption thresholds (Fig. ~\ref{fig1}e-g). Further, prior beliefs moved thresholds in the direction expected if the agent were adopting based on coherence of incoming beliefs with its priors. For example, 22$^\circ$C had a relatively low threshold under a normal prior, but when the model expected a heat wave or a cold snap, the threshold for adopting belief in this temperature was substantially higher (Fig. ~\ref{fig1}e). Prior beliefs about the reliability of the source influenced adoption thresholds of information coming from individuals who were described as having a fixed reliability (Fig. ~\ref{fig1}f). Lastly, prior beliefs about the relative value of personal knowledge versus social input spanned the full range of adoption thresholds for all levels of plausibility considered (Fig. ~\ref{fig1}e). These results provide evidence that coherence between prior and incoming beliefs is driving adoption in LLM agents. Coherence is determined by the effective distance between a prior belief and the new one and can be manipulated by installing a particular prior or by modifying each of the three axes of claim plausibilit, source reliability or agent persona.

\subsection{Reduction to a one-dimensional effective coherence}
\begin{figure}[t]
  \centering
  \includegraphics[width=\linewidth]{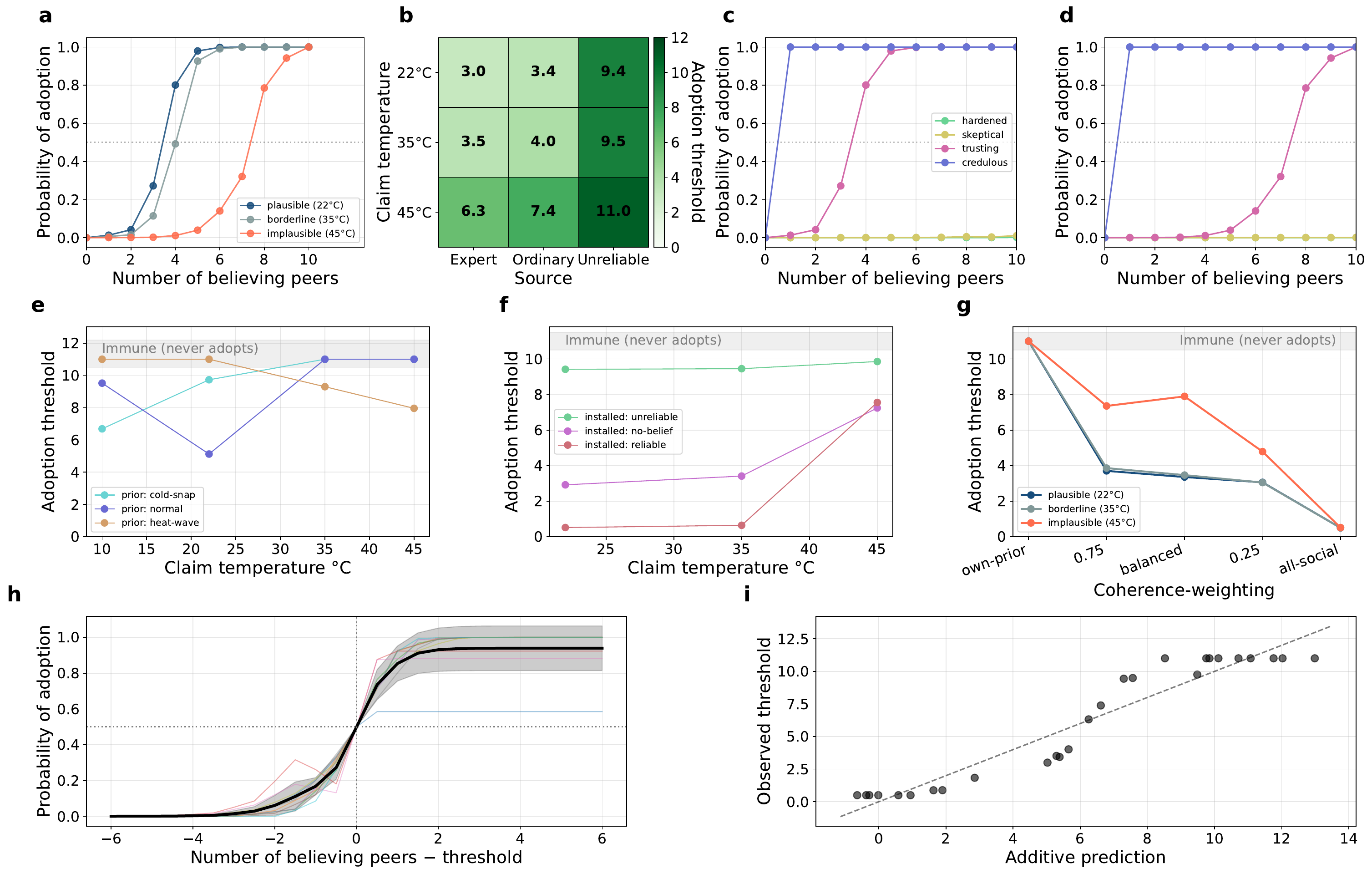}
 \caption{\textbf{Belief adoption is coherence-driven and reduces to one effective dimension.}
  \textbf{(a)} Adoption kernels $p_{\mathrm{adopt}}(k)$ for plausible / borderline /
  implausible claims; less plausible claims require more endorsing peers, raising the
  threshold. \textbf{(b)} Adoption threshold across claim plausibility and source
  reliability: expert sources lower the threshold and compensate for implausibility, while
  an unreliable source raises it. \textbf{(c, d)} Kernels for four personas at two
  plausibility levels, tracing the resilience ladder from hardened (ignores peers) through
  trusting to credulous (adopts at minimal pressure). \textbf{(e-g)} The three axes are
  modifiable coherence relations: installing \textbf{(e)}~a world prior, \textbf{(f)}~a
  source-reliability belief, or \textbf{(g)}~a coherence-weighting instruction moves the
  threshold in the direction a coherence account predicts. \textbf{(h)}~Kernels shifted by
  their own threshold collapse onto a single master curve (individual conditions in colour,
  mean in black), and \textbf{(i)}~an additive model predicts the threshold across the
  $3{\times}3{\times}3$ grid. The threshold tensor is one-dimensional (first PC
  \textbf{98\%}) and additive (\textbf{92\%}).}
  \label{fig1}
\end{figure}

That adoption is coherence-driven raises the question of how the three axes are represented
internally. Three results indicated that the LLM collapses them onto a single dimension. First,
kernels generated by manipulating different sources of coherence, once shifted by their own
thresholds, collapsed onto a single master curve of common shape (Fig.~\ref{fig1}h). Second,
applying dimensionality reduction to the threshold tensor $\Theta$, the first principal component
accounted for \textbf{98\%} of the variance in thresholds. Third, a main-effects model with no
interaction terms achieved $R^2 = \textbf{92\%}$ (Fig.~\ref{fig1}i). Together, these indicate that
the three nominally distinct axes act through one effective, coherence-based dimension.

\section{Population dynamics of belief adoption}
\label{sec:partII}
Having established that belief adoption is well approximated by a single effective coherence dimension, we
placed agents on a network, let them influence one another, and characterized their collective
dynamics. We swept claim
plausibility as a proxy for the single coherence dimension and record the final fraction of believing agents. A belief
is considered to have \emph{ignited} if this fraction exceeds $0.5$.

\subsection{The cascade window and its bifurcation}
\label{sec:bifurcation}
When we swept coherence via claim plausibility, the population underwent a sharp transition from full
adoption to low adoption: ignition probability dropped steeply through a saddle-node-like
transition that closed the cascade window (Fig.~\ref{fig2}a). Away from the transition the
population was monostable, within it the outcome distribution was \emph{bimodal}, so the same
claim under identical conditions produced qualitatively different collective outcomes across
replicates (Fig.~\ref{fig2}b). The steep drop in ignition probability together with these
bimodal outcomes is the signature of an underlying bistability. 

\subsection{Complex contagion: cascade size on clustered networks}
\label{sec:clustering}

We showed that the individual adoption kernel is characteristic of complex contagion (Fig.~\ref{fig1}). Such contagions are expected to produce larger cascades on clustered networks than on random networks of equal density \citep{centola2010,centola2007}. This
advantage is not universal, but it holds in precisely our regime: when adoption without social
reinforcement does not occur (0 in our case) and adoption with reinforcement is very probable
(1 in our case). We tested this on Watts-Strogatz networks of fixed mean degree, varying
only the rewiring probability so the clustering coefficient fell from 
$C{=}\textbf{0.60}$ (ring lattice) to $C{=}\textbf{0.20}$ (essentially random). In
the marginal regime, clustered networks produced significantly larger cascades than random
networks of the same degree (mean final size \textbf{0.67} vs.\ \textbf{0.43}; one-sided
Mann-Whitney $U$, $p=\textbf{2.7e{-}6}$, rank-biserial effect \textbf{0.37}) (Fig.~\ref{fig2}c).

\subsection{Hysteresis: consensus is self-sustaining}
\label{sec:hysteresis}
\begin{figure}[t]
  \centering
  \includegraphics[width=\linewidth]{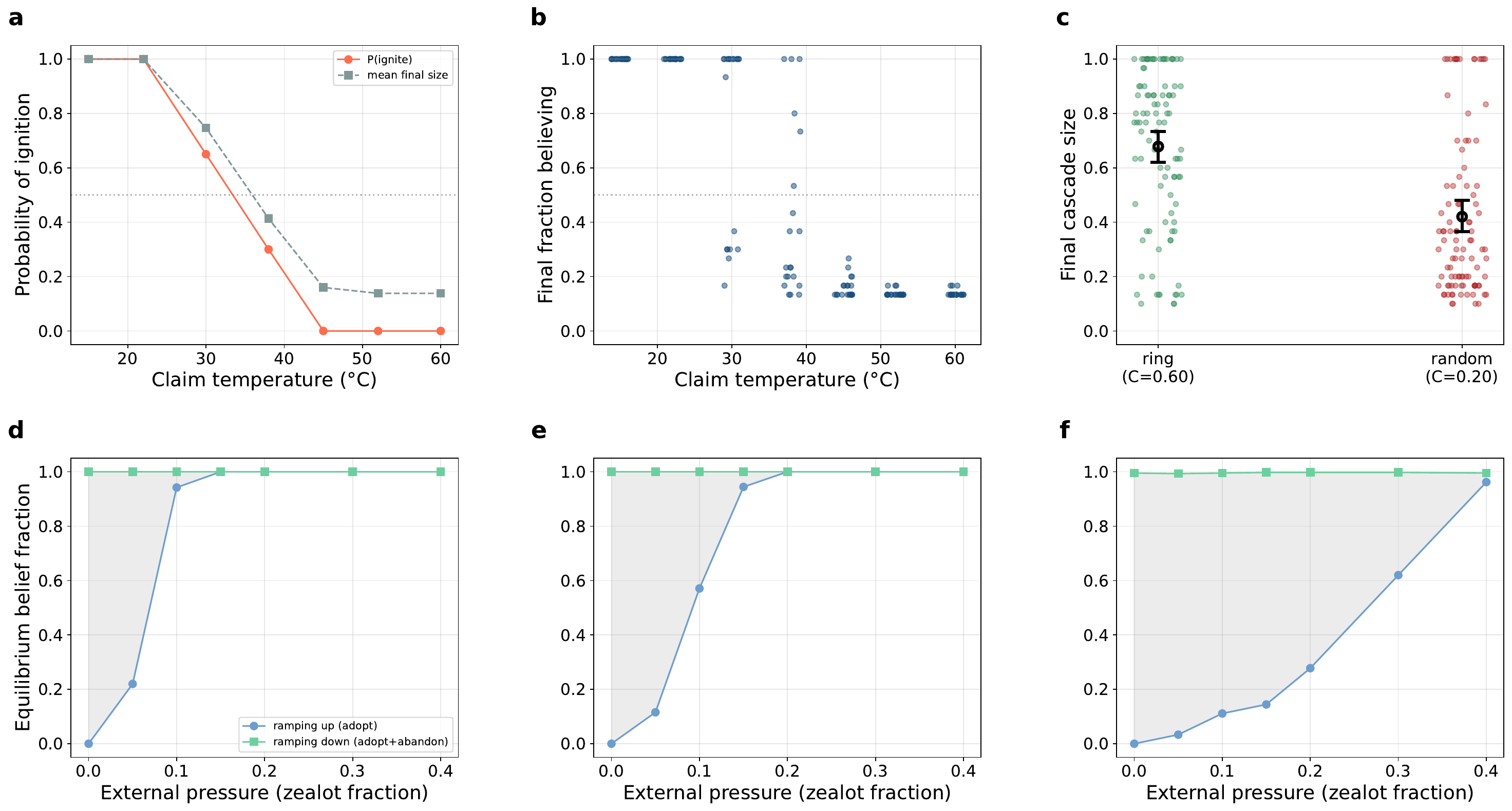}
 \caption{\textbf{Population dynamics of the measured kernel.}
  \textbf{(a)} Ignition probability and mean final cascade size vs.\ claim plausibilit; the
  cascade window closes through a saddle-node-like transition as plausibility falls.
  \textbf{(b)} Per-condition final belief fraction: outcomes are monostable on either side
  of the transition and bimodal within it. \textbf{(c)} Complex contagion: at matched
  degree, clustered networks (ring, $C{=}0.60$) produce larger cascades than random
  networks ($C{=}0.20$), points are individual runs, black shows the mean with $95\%$
  bootstrap CI \textbf{(d-f)}
  Hysteresis for plausible, borderline, and implausible claims: 
  Consensus is self-sustaining; the down-path stays locked at full adoption and the loop
  (shaded) widens with implausibility.}
  \label{fig2}
\end{figure}

A bistable system exhibits hysteresis meaning its state depends on how it was reached. We ramped an
external pressure, the fraction of permanent believers (\emph{zealots}) up from zero and back
down, with agents free to adopt (via the adoption kernel) and abandon (via the retention kernel)
at each step (Fig.~\ref{fig2}d-f). The up-path required substantial pressure to ignite, but once
the population reached consensus the down-path remained locked at full adoption even as all
pressure was removed. The loop widened with implausibility (area \textbf{0.067}, \textbf{0.096},
\textbf{0.252} for plausible, borderline, and implausible claims), because the gap between the
ignition threshold and near-total retention is largest for the least plausible claims. The safety
implication is direct: a false belief is far harder to establish than to maintain, persisting with
no continued support once it reaches consensus. This lock-in is a genuine property of the
agent-driven dynamics, arising from the measured retention kernel rather than a mean-field
artifact.

\section{Robustness across models and domains}
\label{sec:robustness}
Our results so far used one model on one domain. To establish robustness, we replicated along
two independent axes: a second model (\texttt{Mistral-7B-Instruct-v0.3}, temperature domain)
and a second domain (next-day commute time, same model). Every result of
\S\ref{sec:partI}--\ref{sec:partII} reproduced in both; Table~\ref{tab:robustness} summarizes
the central quantities, with full results in Apps. ~\ref{app:mistral}
and~\ref{app:traffic}.

The general structure of the results transferred  without exception. In every condition the adoption kernel was sigmoidal with the same plausibility ordering, the three axes were modifiable coherence
relations, and the threshold tensor was one-dimensional and additive. At the population level,
all three conditions showed the cascade bifurcation with bimodal outcomes, the complex-contagion
clustering signature, and self-sustaining consensus with hysteresis loops that widened with
implausibility.

What varied was calibration, not mechanism. For example, Mistral tended to weight its own priors over social pressure more than Qwen, manifesting as a wider plausibility range, a sharper bifurcation, and wider hysteresis loops. Beliefs in the traffic domain tracked the temperature results closely.


\begin{table}[t]
  \centering
  \small
  \caption{\textbf{Central quantities across the primary condition and two robustness
  axes.} The reduction (dimensionality), the clustering contrast, and the hysteresis
  loop ordering reproduce in every condition.}
  \label{tab:robustness}
  \begin{tabular}{lccc}
    \toprule
     & \textbf{Qwen / temp.} & \textbf{Mistral / temp.} & \textbf{Qwen / traffic}  \\
    \midrule
    Reduction: 1st PC (\%)            & \textbf{98} & \textbf{96} & \textbf{97} \\
    Reduction: additivity $R^2$ (\%)  & \textbf{92} & \textbf{80} & \textbf{91}   \\
    \midrule
    Clustering: $p$ (MW $U$)          & \textbf{2.7e{-}6} & \textbf{4.1e{-}3} & \textbf{7.1e{-}3} \\
    Clustering: effect                & \textbf{+0.37} & \textbf{+0.21} & \textbf{+0.20} \\
    \midrule
    Hysteresis loop (plaus.)          & \textbf{0.067} & \textbf{0.000} & \textbf{0.065} \\
    Hysteresis loop (implaus.)        & \textbf{0.252} & \textbf{0.248} & \textbf{0.248} \\
    \bottomrule
  \end{tabular}
\end{table}

\section{Discussion}
\label{sec:discussion}

We show that LLM agents are susceptible to contagion, with adoption kernels that take the
sigmoidal form underlying complex contagion. These kernels are sensitive to
claim plausibilit, source reliability, agent persona, and the agent's prior beliefs. When
probed, these factors appear to act on the model through a single dimension of effective
coherence between its priors and the incoming belief. At the collective level, beliefs spread
further on clustered networks, and the population exhibits a cascade bifurcation and hysteresis:
once a belief is established, it is harder to dislodge than it was to prevent.

Our results have implications for the safety and controllability of multi-agent systems.
On one hand, a consensus reached by a population of agents (even a false one) may be hard
to correct once established. On the other, belief adoption in naive agents is relatively easy to
steer: installing a simple prior tunes an agent across the full behavioral range, from
readily accepting social input to never accepting it. Such prior installation could protect agents against adopting misinformation.

Finally, our work speaks to the mechanistic basis of these behaviors. A growing literature on the
\emph{linear representation hypothesis} holds that LLMs encode many high-level semantic concepts
as linear directions in their internal representation space \citep{park2024}. Our reduction is a
behavioral analogue: three factors that are conceptually distinct and independently manipulated
collapse onto a single effective dimension of adoption behavior. While we measure
behavior rather than internal representations, this one-dimensionality is consistent with, and
raises the question of whether, the model's internal basis for belief adoption is itself
lower-dimensional than the space of factors that influence it.

\subsection{Limitations and Future Directions}
\label{sec:conclusion}

Our work characterizes coherence in behavior, not in the model's internal representations, a limitation that also points to future work. In non-LLM agents, a model of belief dynamics built on signed, weighted belief networks reproduces both simple- and complex-contagion dynamics, depending on how well an incoming belief coheres with an agent's priors \citep{aiyappa2024}. Installing LLM agents with explicit networks of prior beliefs could connect our behavioral findings to an internal mechanism of belief adoption.

We also consider a single belief spreading in isolation, with no competing claims. A natural
extension is competing contagions, and the question of how to suppress an undesirable belief
while allowing a desirable one to spread. Finally, our populations are small ($N{=}30$) and our
networks stylized, and we study only two models and two belief domains; this separates model
effects from domain effects but cannot establish the pattern across scales or model families.
Because our claims are unverifiable predictions, the results may also not carry over to factual,
morally charged, or multi-claim beliefs, where fact retrieval and cross-claim consistency would
introduce dynamics absent here.


\bibliographystyle{unsrtnat}   
\bibliography{references}





\clearpage
\appendix
\counterwithin{figure}{section}

\section{Full Temperature domain with Mistral results}
\label{app:mistral}

This appendix reports the complete replication of the pipeline of
\S\ref{sec:partI}-\ref{sec:partII} on \texttt{Mistral-7B-Instruct-v0.3}, holding the
temperature domain and all prompts fixed. Every result reproduces qualitatively, with quantitative differences from the primary Qwen results tracing to a
single property: Mistral weights its priors and sources more categorically than
Qwen. This manifests consistently as a wider plausibility range, an all-or-nothing
source axis, 
a sharper cascade bifurcation, and wider hysteresis loops while leaving the mechanisms
themselves intact. Decision tokens were recomputed and verified for Mistral, and the
persona was folded into the user turn, as Mistral does not accept a system role
(\S\ref{sec:method-kernel}, App.~\ref{app:repro}).

\subsection{Individual belief adoption}

\begin{figure}[t]
  \centering
  \includegraphics[width=\linewidth]{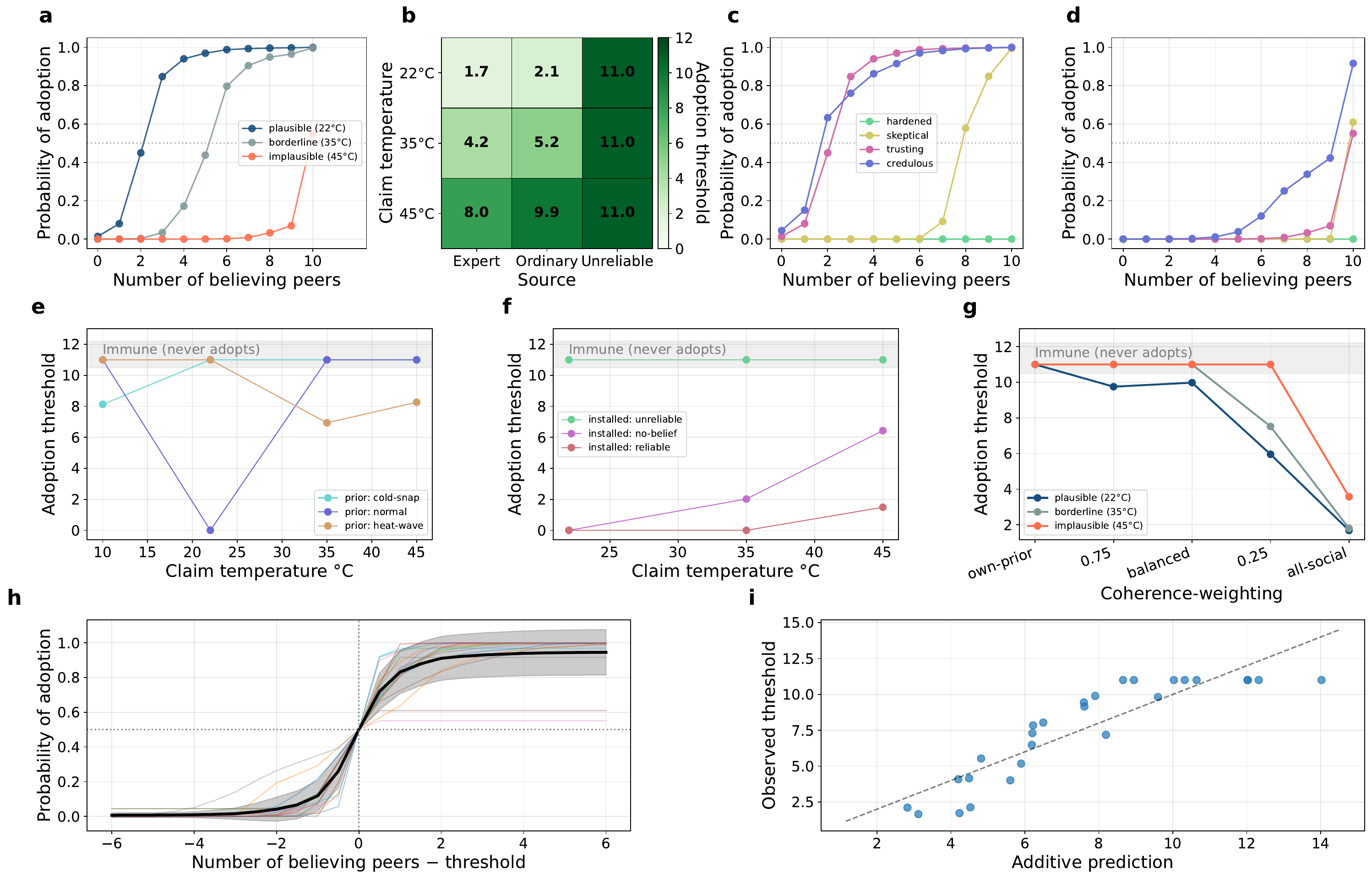}
\caption{\textbf{Mistral: Belief adoption is coherence-driven and reduces to one effective dimension.}
  \textbf{(a)} Adoption kernels for plausible / borderline / implausible claims; the
  threshold rises with implausibility. \textbf{(b)} Adoption threshold across claim
  plausibility and source reliability; expert sources lower the threshold, but an unreliable
  source pins it at the ceiling for every claim, a more categorical source axis than Qwen.
  \textbf{(c)} Kernels per persona at a fixed plausibility level and \textbf{(d)} the same
  under a stronger claim, tracing the resilience ladder from hardened (never adopts) to
  credulous. \textbf{(e-g)} The three axes as editable coherence relations: installing
  \textbf{(e)} a world prior (temperature), \textbf{(f)} a source-reliability belief, and
  \textbf{(g)} a coherence-weighting instruction each moves the threshold in the
  coherence-predicted direction. \textbf{(h)} Kernels shifted by their own threshold
  collapse onto one master curve (individual conditions in colour, mean in black).
  \textbf{(i)} Additive prediction vs.\ observed threshold across the $3{\times}3{\times}3$
  grid. The threshold tensor is one-dimensional (first PC \textbf{96\%}) and additive
  (\textbf{80\%})}
  \label{figS1}
\end{figure}

\paragraph{The three axes (Fig.~\ref{figS1} panels a-d).}
The adoption kernel is sigmoidal with the same plausibility ordering as Qwen, but with a
\emph{wider} dynamic range: thresholds of \textbf{2.1}, \textbf{5.2}, and
\textbf{9.9} peers for plausible, borderline, and implausible claims, versus Qwen's
narrower spread. Mistral adopts plausible claims more readily yet resists implausible
ones more strongly. Source authority again compensates for implausibility and orders
experts $<$ ordinary $<$ unreliable at every claim; unlike Qwen, an unreliable source
produces near-total rejection (threshold at the ceiling) even for plausible claims,
reflecting Mistral's more categorical weighting of source reliability. The resilience
ladder (immune $\to$ sensitive $\to$ credulous) is preserved, with one model-specific
feature: even the credulous persona retains a prior-based floor on the implausible
claim, so plausibility discriminates across a wider range of dispositions than for Qwen.

\paragraph{Coherence tests (Fig.~\ref{figS1} panels e-g).}
All three axes are modifiable coherence relations, and the manipulations are
\emph{sharper} than for Qwen. Installing a world prior moves each claim's threshold to a
near-perfect coherence diagonal: each claim adopts only under the prior that makes it
cohere and resists completely otherwise. Installing a source-reliability belief swings
the threshold from the ceiling (unreliable, never adopts) to near zero (reliable, adopts
with minimal pressure), a swing of up to \textbf{8.9} peers from editing a single
belief. Instructing the coherence-weighting reproduces the immune-sensitive-credulous
ladder with no personality descriptors, though the descent is steeper and later than
Qwen's, consistent with Mistral's stronger default prior-weighting.

\paragraph{The reduction (Fig.~\ref{figS1} panels h-i).}
The central result holds: the threshold tensor is one-dimensional, with the first
principal component capturing \textbf{96\%} of the variance. The additive model
explains \textbf{80\%} of the variance, lower than Qwen's \textbf{92\%},
because Mistral's more categorical (thresholds pinned at the ceiling or near zero)
behavior introduces more interaction. The one-dimensional reduction is thus model-invariant.

\subsection{Population dynamics of belief adoption}

\begin{figure}[t]
  \centering
  \includegraphics[width=\linewidth]{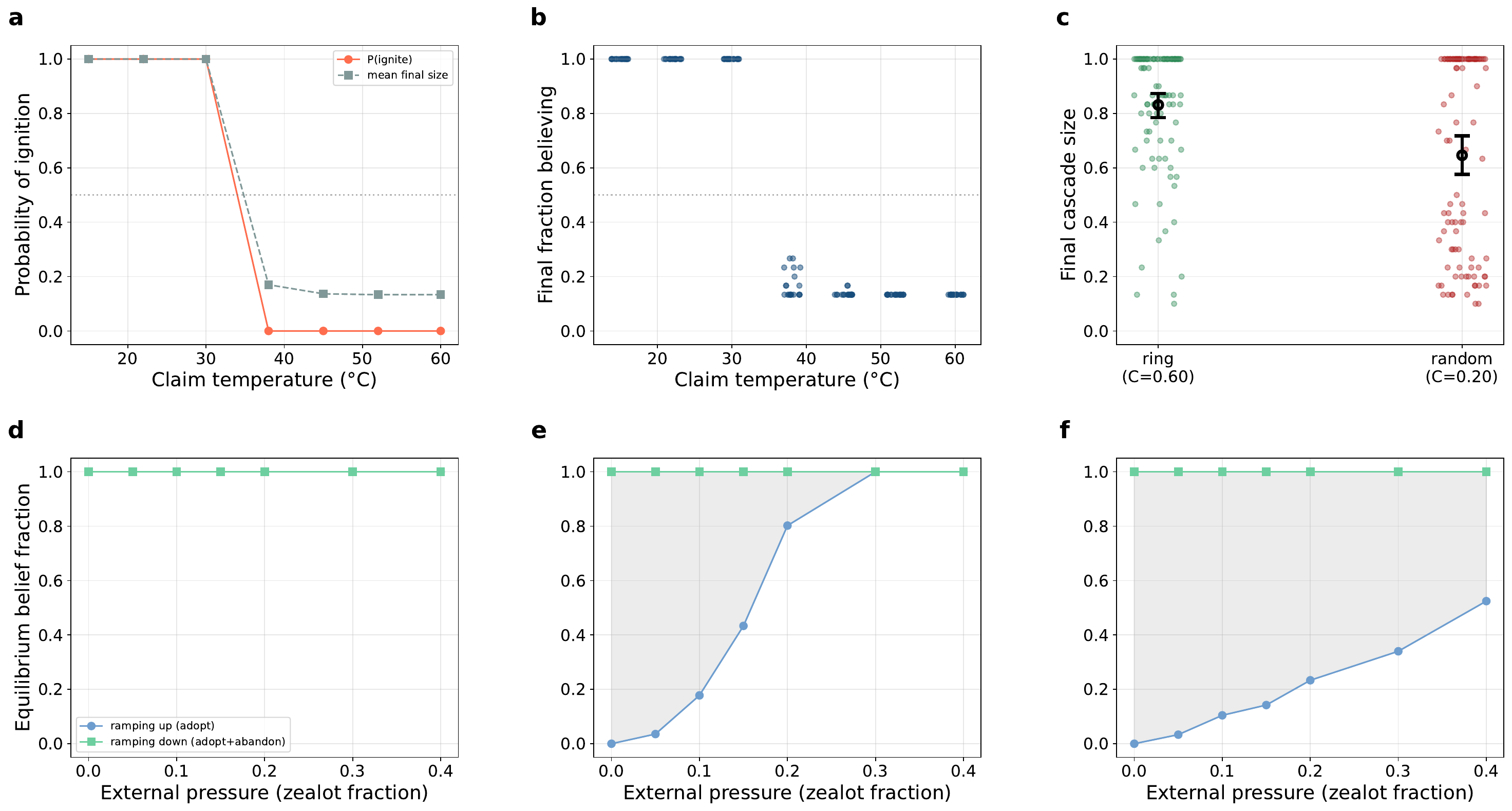}
  \caption{\textbf{Population dynamics of the measured kernel.}
  \textbf{(a)} Ignition probability and mean final cascade size vs.\ claim plausibilit; the
  window closes through a sharp bifurcation. \textbf{(b)} Per-condition final belief
  fraction, monostable on either side and bimodal within the narrow transition zone.
  \textbf{(c)} Complex contagion: at matched degree, clustered (ring, $C{=}0.60$) networks
  produce larger cascades than random ($C{=}0.20$) networks (points are individual runs;
  black shows mean with $95\%$ bootstrap CI)
  \textbf{(d-f)} Hysteresis for plausible, borderline, and implausible claims. Consensus is self-sustaining and the loop (shaded) widens with implausibility.}
  \label{figS2}
\end{figure}

\paragraph{Bifurcation (Fig.~\ref{figS2} panels a-b).}
The cascade window closes through a bifurcation, sharper than Qwen's: ignition
probability drops from one to zero over a narrow plausibility interval, with the
transition at \textbf{$\sim$34$^\circ$C}. The outcome distribution is monostable on
either side and bimodal within the (correspondingly narrow) transition zone. Mistral's
sharper transition leaves a narrower marginal regime than Qwen's, an observation that
matters for the clustering experiment below.

\paragraph{Clustering (Fig.~\ref{figS2} panel c).}
The complex-contagion signature holds: at matched degree, clustered networks produce
larger cascades than random networks (mean final size \textbf{0.83} vs.\
\textbf{0.65}; one-sided Mann-Whitney $U$, $p = \textbf{4.1e{-}3}$, rank-biserial
effect \textbf{+0.21}). The effect is moderate and
smaller than Qwen's, because Mistral's higher baseline ignition in the marginal regime
compresses the headroom for topology to discriminate, consistent with, and explained
by, its sharper bifurcation.

\paragraph{Hysteresis (Fig.~\ref{figS2} panels d-f).}
Consensus is self-sustaining and the down-path remains locked at full adoption as
pressure is removed. The hysteresis loop widens with implausibility (loop area
\textbf{0.000}, \textbf{0.159}, \textbf{0.248} for plausible, borderline, and
implausible), wider than Qwen's for the borderline claim and comparable for the implausible one: Mistral
resists establishing an implausible belief more strongly on the up-path while consensus
locks equally firmly on the down-path, so the up-down gap is larger. The plausible loop
is near zero because Mistral adopts plausible claims so readily that the up-path itself
saturates immediately.

\paragraph{Summary.}
Every result reproduces on Mistral. The mechanisms: coherence-driven resistance, its
reduction to one effective scalar, the cascade bifurcation, the clustering signature,
and self-sustaining consensus, are model-invariant; the magnitudes and the identity of
the dominant resistance axis are model-specific and are all accounted for by Mistral's
stronger, more categorical weighting of priors and sources.

\clearpage
\section{Full traffic-domain with Qwen results}
\label{app:traffic}

This appendix reports the complete replication of the pipeline of
\S\ref{sec:partI}-\ref{sec:partII} on a second claim domain, holding the model fixed as
\texttt{Qwen2.5-3B-Instruct}. Claims are next-day commute times of the form ``Tomorrow
morning's commute on I-93 into Boston will take about \textit{[duration]}'' with
plausibility set by the stated duration, an unverifiable prediction with a graded prior,
like temperature, but with a softer ceiling (long delays remain conceivable, so the
implausible level sits at a multi-hour commute). Every result reproduces.
The traffic results track the temperature results closely, including nearly identical
reduction statistics and hysteresis loop areas, confirming that the mechanism is not an
artifact of the temperature substrate. We map three plausibility levels (a normal, a long,
and an absurd commute) onto durations whose kernels match temperature's plausible /
borderline / implausible thresholds, with an additional impossible value for the source
experiments (App.~\ref{app:repro}).

\subsection{Individual belief adoption}

\begin{figure}[t]
  \centering
  \includegraphics[width=\linewidth]{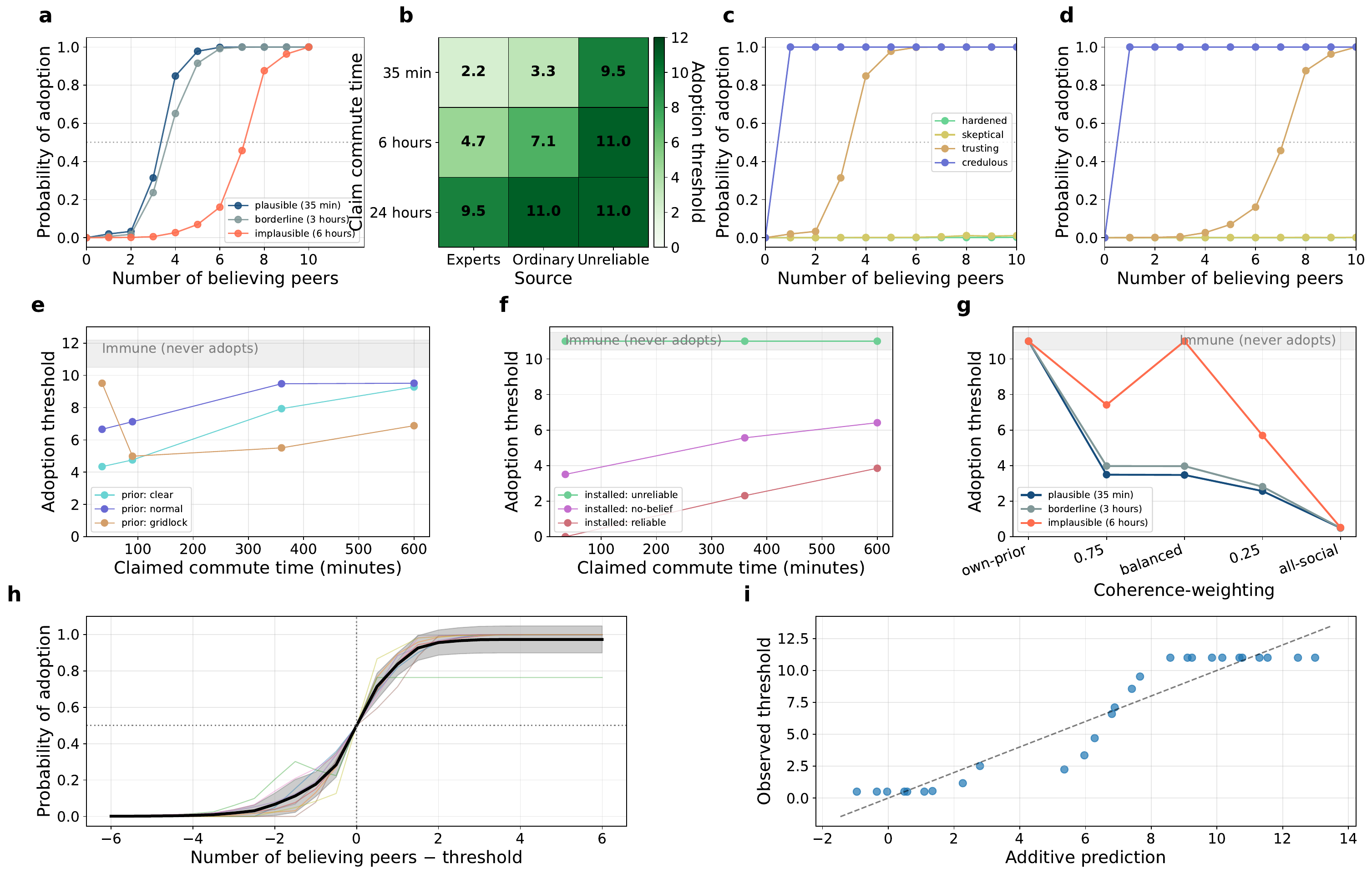}
  \caption{\textbf{Traffic domain, Belief adoption is coherence-driven and reduces to one effective dimension.}
  \textbf{(a)} Adoption kernels for plausible (35 min) / borderline (3 h) / implausible
  (6 h) commute-time claims; the threshold rises with implausibility. \textbf{(b)} Adoption
  threshold across claim plausibility and source reliability; expert sources lower the
  threshold while an unreliable source pins it near the ceiling, reproducing the
  plausibility wall. \textbf{(c)} Kernels per persona at a fixed plausibility level and
  \textbf{(d)} the same under a stronger claim, tracing the resilience ladder from hardened
  (never adopts) to credulous. \textbf{(e-g)} The three axes as editable coherence
  relations: installing \textbf{(e)} a road-condition prior (clear / normal / gridlock),
  \textbf{(f)} a source-reliability belief, and \textbf{(g)} a coherence-weighting
  instruction each moves the threshold as a coherence account predicts; the road-condition
  prior acts \emph{bidirectionally}, lowering the threshold for long commutes and raising it
  for short ones. \textbf{(h)} Kernels shifted by their own threshold collapse onto one
  master curve (individual conditions in colour, mean in black). \textbf{(i)} Additive
  prediction vs.\ observed threshold across the $3{\times}3{\times}3$ grid. The threshold
  tensor is one-dimensional (first PC \textbf{97\%}) and additive (\textbf{91\%}), closely
  matching the temperature primary.}
  \label{figS3}
\end{figure}

\paragraph{The three axes (Fig.~\ref{figS3}) panels a-d.}
The adoption kernel is sigmoidal with the same plausibility ordering: thresholds of
\textbf{3.3}, \textbf{3.6}, and \textbf{7.1} peers for the plausible, borderline,
and implausible commutes, closely matching temperature's plausible/implausible spread.
(The borderline level is softer than temperature's, sitting nearer the plausible end,
because the model finds a wide range of commute durations believable, the one substantial
domain difference.) Source authority compensates for implausibility and orders experts
$<$ ordinary $<$ unreliable at every claim, and an authoritative source moves even the
impossible commute toward adoption, reproducing the plausibility-wall breach seen for
temperature. The resilience ladder (immune $\to$ sensitive $\to$ credulous) is
preserved, with plausibility discriminating only in the sensitive band.

\paragraph{Coherence tests (Fig.~\ref{figS3} panels e-g).}
All three axes are modifiable coherence relations. Installing a road-condition prior (clear, normal, or gridlock) moves a fixed claim's threshold in the coherence-predicted direction: a gridlock prior lowers the threshold for long commutes while raising it for short ones, and vice versa for a clear-roads prior. This provides a second, independent demonstration of the coherence mechanism in a distinct domain, complementing the bidirectional temperature result of \S\ref{sec:partI} (Fig.~\ref{fig1}e). Installing a source-reliability belief swings
the threshold from the ceiling to near zero, and instructing the coherence-weighting
reproduces the immune-sensitive-credulous ladder with no personality descriptors.

\paragraph{The reduction (Fig.~\ref{figS3} panels h-i).}
The central result holds and closely matches the temperature primary: the threshold
tensor is one-dimensional, with the first principal component capturing
\textbf{97\%} of the variance (versus \textbf{98\%} for temperature), and the
additive model explains \textbf{91\%} (versus \textbf{92\%}). 

\subsection{Population dynamics of belief adoption}

\begin{figure}[t]
  \centering
  \includegraphics[width=\linewidth]{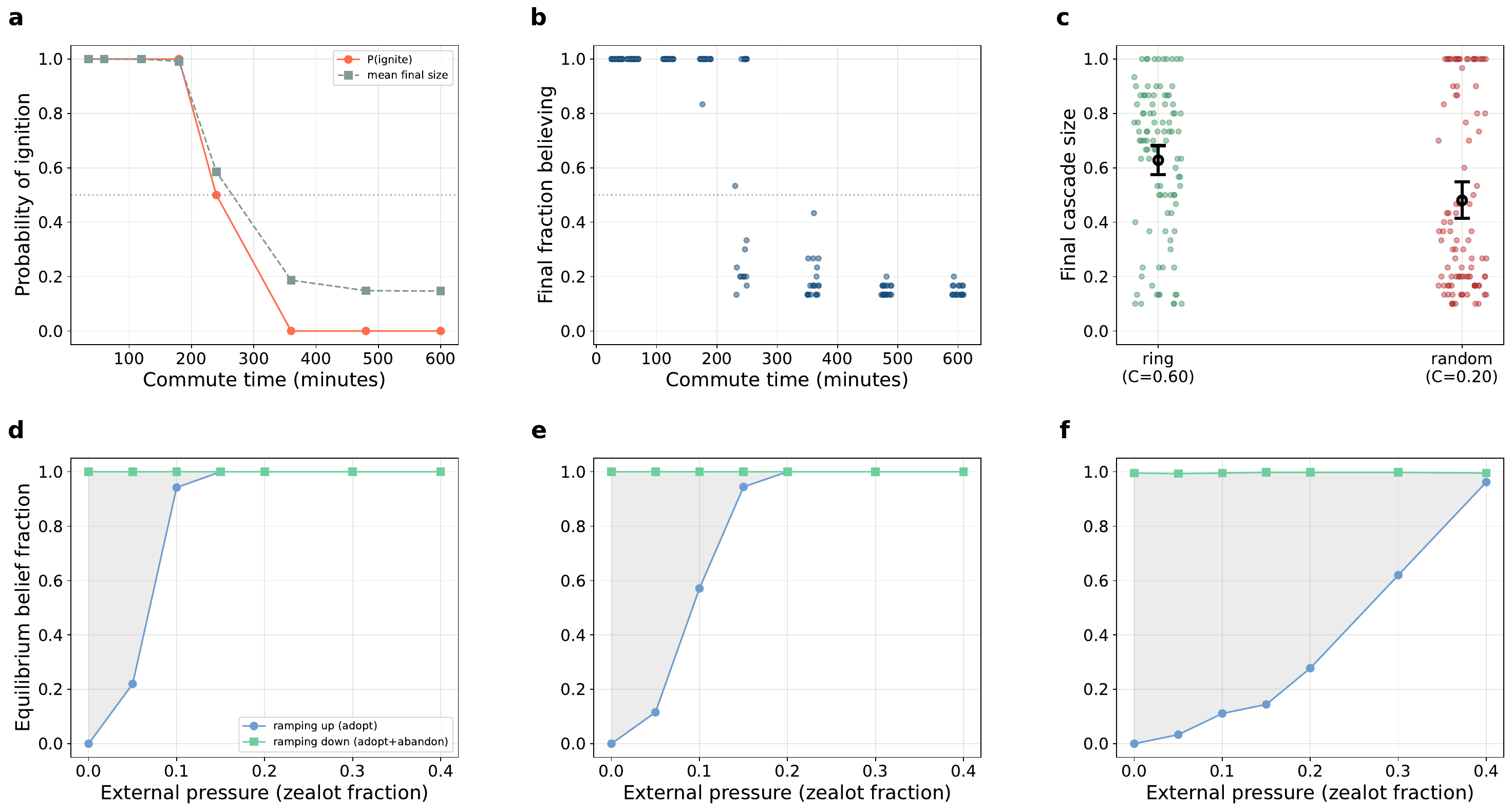}
  \caption{\textbf{Population dynamics of the measured kernel.}
  \textbf{(a)} Ignition probability and mean final cascade size vs.\ claimed commute time;
  the cascade window closes through a clean transition at a commute of $\sim$4 hours.
  \textbf{(b)} Per-condition final belief fraction, monostable on either side and bimodal
  within the transition zone. \textbf{(c)} Complex contagion: at matched degree, clustered
  (ring, $C{=}0.60$) networks produce larger cascades than random ($C{=}0.20$) networks
  (points are individual runs; black shows mean with $95\%$ bootstrap CI) \textbf{(d-f)} Hysteresis for plausible, borderline, and
  implausible claims. Consensus is self-sustaining; the down-path
  stays locked at full adoption and the loop (shaded) widens with implausibility, with
  loop areas nearly identical to the temperature primary.}
  \label{figS4}
\end{figure}

\paragraph{Bifurcation (Fig.~\ref{figS4} panels a-b).}
The cascade window closes through a clean, monotonic transition: ignition probability
falls from one for short commutes to zero for absurd ones, crossing $0.5$ at a commute of
approximately four hours. The transition is if anything cleaner than
temperature's, as the traffic plausibility gradient is smooth over this range. The outcome
distribution is monostable on either side and bimodal within the transition zone,
reproducing the bistability signature.

\paragraph{Clustering (Fig.~\ref{figS4} panel c).}
The complex-contagion signature holds: at matched degree, clustered networks produce
larger cascades than random networks (mean final size \textbf{0.62} vs.\
\textbf{0.48}; one-sided Mann--Whitney $U$, $p = \textbf{7.1e{-}3}$, rank-biserial
effect \textbf{+0.20}), measured in the marginal
regime identified by the bifurcation.

\paragraph{Hysteresis (Fig.~\ref{figS4} panels d-f).}
Consensus is self-sustaining and the down-path remains locked at full adoption as
pressure is removed. The hysteresis loop widens with implausibility (loop area
\textbf{0.065}, \textbf{0.080}, \textbf{0.248} for plausible, borderline, and
implausible), \emph{nearly identical} to the temperature loop areas
(\textbf{0.067}, \textbf{0.096}, \textbf{0.252}).

\paragraph{Summary.}
Every result reproduces on the traffic domain, with quantities that closely match the
temperature primary throughout. Together with the model replication of
App.~\ref{app:mistral}, this establishes robustness along both axes: the model shifts
magnitudes and the identity of the dominant resistance axis, the domain shifts almost
nothing, and the mechanisms are invariant under both.


\clearpage
\section{Reproducibility details}
\label{app:repro}

This appendix specifies all prompts, tokens, mappings, and parameters needed to
reproduce our results. All measurements use a single, sampling-free forward pass in half precision: we read the decision logits directly rather than generating tokens. Every reported quantity is deterministic given the prompt and the random seed (which fixes graph generation, initial seeding, and all Bernoulli adoption/abandonment draws).
The models we use (Qwen2.5-3B-Instruct and Mistral-7B-Instruct-v0.3) are released under the
Apache~2.0 license. Code and configuration are available at
\url{https://anonymous.4open.science/r/multi-agent-llm-contagion-F402}.

\subsection{Prompt templates}
\label{app:repro-prompts}
Prompts consist of a system prompt (the persona) and a user prompt. For models without a
system role (e.g.\ Mistral), the persona is prepended to the user prompt, separated by a
blank line. Placeholders are written in \texttt{\{braces\}}.

\paragraph{Canonical persona (system prompt).}
\begin{quote}\small\ttfamily
You are Morgan, a reasonable person. You weigh what your peers think. If enough trusted
peers accept a claim, you tend to accept it too, unless it is clearly false.
\end{quote}

\paragraph{Adoption kernel (user prompt).}
\begin{quote}\small\ttfamily
\{prior\}Claim: ``\{claim\}''\\[2pt]
\{social\}\\[2pt]
Do you accept this claim? Answer with exactly one word: ADOPT or DECLINE.
\end{quote}
The \texttt{\{social\}} line is, for $k>0$,
``\texttt{\{k\} of your 10 peers \{who\} have already accepted this as true; the other
\{10-k\} have not weighed in.}'', and for $k=0$,
``\texttt{None of your 10 peers have expressed a view on this.}'' The optional
\texttt{\{who\}} inserts a source descriptor (``\texttt{--- \{source\} ---}'') in the
source experiments and is empty otherwise. The optional \texttt{\{prior\}} prepends an
installed belief (below) in the coherence experiments and is empty otherwise. At $k{=}10$ this template produces the degenerate phrasing ``\texttt{\dots the other 0 have
not weighed in.}'' This does not distort the measured kernels: $p_{\mathrm{adopt}}(10)$ lies
on the smooth continuation of the kernel through $k{=}9$ in every condition, with no
discontinuity at the endpoint. For most conditions the $0.5$ crossing falls well below
$k{=}9$, so the threshold does not depend on this point; for the few whose crossing falls
between $k{=}9$ and $k{=}10$, the interpolated threshold does depend on it, but its value is
consistent with the rest of the kernel and with the ordering across conditions. For ceiling
conditions ($\theta{=}11$), the threshold reflects only that $p_{\mathrm{adopt}}(10)<0.5$,
adoption remains a minority even at full peer count,not its exact value, and is therefore
robust to the endpoint phrasing. The retention template special-cases $k{=}10$ because
retention is evaluated at full consensus, where that endpoint is load-bearing throughout.

\paragraph{Retention kernel (user prompt).}
Used for the reversible (hysteresis) dynamics, for an agent that already holds the belief:
\begin{quote}\small\ttfamily
You currently believe this claim: ``\{claim\}''\\[2pt]
\{social\}\\[2pt]
Do you maintain this belief? Answer with exactly one word: MAINTAIN or ABANDON.
\end{quote}
Here \texttt{\{social\}} reports how many peers \emph{still} accept the claim: for
$0<k<10$, ``\texttt{\{k\} of your 10 peers still accept this as true; the other \{10-k\}
have dropped it.}''; for $k=10$, ``\texttt{All 10 of your peers still accept this as
true.}''; for $k=0$, ``\texttt{None of your 10 peers accept this anymore; all have
dropped it.}''

\paragraph{Coherence manipulations.}
Each installs a belief via the \texttt{\{prior\}} slot (world prior, source reliability)
or by replacing the persona (coherence weighting). Representative installed beliefs, with
the level label used in the figures given in brackets:
\begin{itemize}\small
\item \textbf{Claim prior (temperature).}
``\texttt{You know that Boston has been in a record heat wave this week, with
temperatures well above normal, around 40\textdegree{}C (104\textdegree{}F).}''
(analogously, [cold-snap] and [normal] statements).
\item \textbf{Claim prior (traffic).}
``\texttt{You know that a major snowstorm has hit Boston, roads are severely backed up,
and there are multiple accidents causing hours-long delays.}''
(analogously, [clear] and [normal] statements).
\item \textbf{Source prior.}
``\texttt{You know that these \{source\} have an excellent track record and are almost
always right.}'' (analogously, [unreliable] and [no-belief]).
\item \textbf{Disposition prior.}
``\texttt{You are a person deciding whether to adopt beliefs. You weigh your own prior
knowledge and judgment very heavily. You almost never adopt a claim based on peer opinion
alone; peers barely move you.}'' (graded through [balanced] to [all-social]:
``\texttt{\dots\ You weigh peer consensus almost entirely\dots}'').
\end{itemize}

\paragraph{Personas (resilience experiments).}
The resilience ladder uses four system prompts: \emph{hardened} (``\texttt{highly
skeptical and independent\dots\ peer opinion barely affects you}''), \emph{skeptical}
(``\texttt{cautious\dots\ need very strong social consensus}''), \emph{trusting} (the
canonical persona above), and \emph{credulous} (``\texttt{highly socially attuned and
trusting\dots\ you readily accept it too}''). Full strings are in the released config.

\subsection{Decision tokens}
\label{app:repro-tokens}

We read the logits of the first sub-word token of each decision word, recomputed per
model. Table~\ref{tab:tokens} lists the token identifiers; each pair was verified to be
distinct and to match greedy decoding on a clear prompt.

\begin{table}[h]
  \centering\small
  \caption{First sub-word token identifiers per model.}
  \label{tab:tokens}
  \begin{tabular}{llcc}
    \toprule
    Model & Decision & Token & ID \\
    \midrule
   \multirow{4}{*}{\texttt{Qwen2.5-3B-Instruct}}
  & ADOPT    & \texttt{ADO}  & 28814 \\
  & DECLINE  & \texttt{DECL} & 21817 \\
  & MAINTAIN & \texttt{MA}   & 4835 \\
  & ABANDON  & \texttt{AB}   & 1867 \\
    \midrule
    \multirow{4}{*}{\texttt{Mistral-7B-Instruct-v0.3}}
      & ADOPT    & \texttt{AD} & 10772 \\
      & DECLINE  & \texttt{DE} & 5330 \\
      & MAINTAIN & \texttt{MA} & 7308 \\
      & ABANDON  & \texttt{AB} & 19090 \\
    \bottomrule
  \end{tabular}
\end{table}

\subsection{plausibility mappings}
\label{app:repro-mappings}

Table~\ref{tab:mappings} gives the specific claim values used for each plausibility level
in each domain, chosen so that the plausible and implausible kernels are well separated;
the impossible value is used only in the source experiments (plausibility-wall test).

\begin{table}[h]
  \centering\small
  \caption{Claim values by plausibility level and domain.}
  \label{tab:mappings}
  \begin{tabular}{lll}
    \toprule
    plausibility level & Temperature (Boston) & Traffic (I-93 commute) \\
    \midrule
    plausible    & \textbf{22$^\circ$C}  & \textbf{35 min} \\
    borderline   & \textbf{35$^\circ$C}  & \textbf{3 hours} \\
    implausible  & \textbf{45$^\circ$C}  & \textbf{6 hours} \\
    impossible   & \textbf{75$^\circ$C}  & \textbf{24 hours} \\
    \bottomrule
  \end{tabular}
\end{table}

\subsection{Source and reliability descriptors}
\label{app:repro-sources}

Source experiments use three descriptors: \emph{expert}
(``\texttt{professional meteorologists}'' for temperature; ``\texttt{traffic authorities
and the state transportation department}'' for traffic), \emph{ordinary} (no descriptor),
and \emph{unreliable} (``\texttt{people known for spreading rumors}''). The
source-reliability coherence test (\S\ref{sec:method-substrate}) holds the source label
fixed (``\texttt{professional meteorologists}'' / ``\texttt{traffic reporters}'') and
varies only the installed reliability belief.

\subsection{Population and network parameters}
\label{app:repro-params}

Please see Table \ref{tab:params}.

\begin{table}[h]
  \centering\small
  \caption{Population simulation parameters, per experiment. Each quantity is fixed (not a
  range); stochasticity across replicates comes only from the random seed, enumerated
  $0,\dots,R-1$.}
  \label{tab:params}
  \resizebox{\textwidth}{!}{%
  \begin{tabular}{llll}
    \toprule
    Parameter & Bifurcation (\S\ref{sec:bifurcation}) & Clustering (\S\ref{sec:clustering}) & Hysteresis (\S\ref{sec:hysteresis}) \\
    \midrule
    Network size $N$        & 30 & 30 & 30 \\
    Peer discretization $n$ & 10 & 10 & 10 \\
    Topology                & scale-free (BA, $m{=}3$) & Watts--Strogatz (deg.\ 6) & scale-free (BA, $m{=}3$) \\
    Seed fraction $\rho_0$  & 0.15 & 0.12 & --- \\
    Rewiring $p$            & --- & $\{0\ (\text{ring}),\,1\ (\text{random})\}$ & --- \\
    Horizon (steps)         & 15 & 15 & 20 \\
    Replicates $R$          & 20 & 100 & 15 \\
    plausibility             & $T\in\{15,22,30,38,45,52,60\}^\circ$C & $T{=}36^\circ$C & $T\in\{22,35,45\}^\circ$C \\
    Zealot ramp             & --- & --- & $\{0,0.05,0.10,0.15,0.20,0.30,0.40\}$ \\
    \bottomrule
  \end{tabular}}
\end{table}

\subsection{Statistical procedures}
\label{app:repro-stats}

Bootstrap confidence intervals use \textbf{5000} resamples of the per-condition final
sizes; reported intervals are the $2.5$ and $97.5$ percentiles of the resampled means.
Two-group contrasts (clustered vs.\ random) use the one-sided Mann-Whitney $U$ test
(\texttt{scipy.stats.mannwhitneyu}, \texttt{alternative=`greater'}); the rank-biserial
correlation effect size is $r = 2U/(n_1 n_2) - 1$. Ignition probability is the fraction of replicates whose final belief
fraction exceeds $0.5$.

\subsection{Compute}
\label{app:repro-compute}

All experiments run on a single NVIDIA L4 GPU (24~GB) in half precision. The full pipeline for one (model, domain) condition, all kernel measurements and population simulations, completes in approximately one GPU-hour; the population simulations dominate, as each cascade step requires one forward pass per distinct believing-neighbor count.


\end{document}